# FPGA-based Acceleration System for Visual Tracking

Ke Song[1, 2], Chun Yuan[2*], Peng Gao[1], Yunxu Sun[1]

[1] Shenzhen Graduate School, Harbin Institute of Technology, Shenzhen 518055, China
[2] Graduate School at Shenzhen, Tsinghua University, Shenzhen 518055, China
* Email: yuanc@sz.tsinghua.edu.cn

**Abstract**

Visual tracking is one of the most important application areas of computer vision. At present, most algorithms are mainly implemented on PCs, and it is difficult to ensure real-time performance when applied in the real scenario. In order to improve the tracking speed and reduce the overall power consumption of the visual tracking, this paper proposes a real-time visual tracking algorithm based on DSST(Discriminative Scale Space Tracking) approach. We implement a hardware system on Xilinx XC7K325T FPGA platform based on our proposed visual tracking algorithm. Our hardware system can run at more than 153 frames per second. In order to reduce the resource occupation, our system adopts the batch processing method in the feature extraction module. In the filter processing module, the FFT IP core is time-division multiplexed. Therefore, our hardware system utilizes LUTs and storage blocks of 33% and 40%, respectively. Test results show that the proposed visual tracking hardware system has excellent performance.

## 1. Introduction

Visual tracking is an important branch in the field of computer vision. tracking on motion platforms uses embedded devices as the processor. Its computing resources and storage resources are limited, and some of them also require the power consumption, volume, weight, etc. In addition, the use of high-resolution cameras also increases the amount of computation of visual tracking, and makes it difficult to handle high frame rates. Field Programmable Gate Array (FPGA) has a parallel processing structure and superior flexibility. It is suitable for the implementation of large-data throughput data interfaces and visual algorithms. At the same time, FPGA has low power consumption and high integrating capability, which can reduce the overall power consumption and volume of the embedded device. Therefore, using FPGA as a hardware development platform has important theoretical significance and application value.

At present, many researchers have studied on the design of visual tracking hardware system. Miao *et al.* [1] completed a dedicated processing chip for high-speed visual tracking which can perform real-time tracking in a simple background. Jung [2] designed a real-time tracking system based on adaptive color histogram on FPGA. The processing frequency can reach 81FPS for 15x15 pixels. Elkhatib [3] proposed a hardware system that uses Altera Cyclone II platform to process a resolution of 640x480 images under 20MHz clock conditions. Its frame rate can reach 21.7 FPS. The system architecture of the SIFT+BRIEF architecture proposed by Wang [4], which shows that the 720p image can reach 60 FPS on FPGA. Tahara [5] proposed a particle filter-based target tracking system using Xilinx Kintex VII FPGA and achieved good results. These systems have achieved excellent results, but all the results were obtained by sacrificing the real-time capabilities. In addition, some FPGAs are relatively expensive to select and are not suitable for general application scenarios. In this paper, we propose a real-time visual tracking hardware system based on Xilinx XC7K325T FPGA platform. Use a lower-cost FPGA and optimize the algorithm to reduce the amount of calculations. In addition, resource optimization and structure optimization are implemented for the hardware implementation to reduce resource consumption, and the accuracy is greatly improved while ensuring accuracy.

## 2. Algorithm Theory and Hardware Implementation

DSST [6] is a tracking algorithm based on correlation filters. It has good robustness to motion blur and illumination changes, and can estimate the target scale. In this paper, 1-D gray feature and 32-dimensional (HOG) feature are selected to estimate the target position. The 32-dimensional HOG(Histogram of Oriented Gradients) feature is used to estimate 7 target dimensions. The two processes are independent, making the algorithm easier to parallelize. The hardware design is divided into seven modules: 1) image extraction module; 2) scale calculation module; 3) interpolation module; 4) HOG feature extraction module; 5) correlation filter calculation and update module; 6) target position and scale calculation module 7) target information update module.

### 2.1 Algorithm Theory

DSST selects multidimensional features which indexed by the feature channel $l \in \{1,2,\ldots,d\}$ during sample extraction. The multidimensional features of the input sample $f$ are composed of grayscale and HOG.

Compared with MOSSE(Minimum Output Sum of Squared Error filter), the increased HOG feature makes the algorithm adapt to the texture feature scene better. The minimum mean square error sum can be expressed as,

$$\varepsilon = \left\| \sum_{l=1}^{d} h^l * f^l - g \right\|^2 + \lambda \sum_{l=1}^{d} \left\| h^l \right\|^2 \quad (1)$$

Where $g$ is the desirable output that supposed to follow Gaussian distribution, and $j$ indicates the current processed frame number. The image size of $h$, $g$, $f$ is $M\times N$. The parameter $\lambda$ is a regular term which can eliminate the influence of the zero component in the spectrum of the sample $f$ and avoid the denominator of the above solution being zero. Based on the circulant properties of correlation filters, the time domain solution can be converted to the frequency domain solution as the following formula,

$$H^l = \frac{\overline{G}F^l}{\sum_{k=1}^{d} \overline{F^k}F^k + \lambda} = \frac{A_t^l}{B_t} \quad (2)$$

Where $l$ is a dimension of the feature, $t$ represents the current frame. However, solving the linear equations of $d \times d$ dimension is time-consuming. In order to obtain a robust approximation, the numerator $A_t^l$ and denominator $B_t$ in the above equation are separately updated as follows,

$$A_t^l = (1-\eta)\ A_{t-1}^l + \eta \overline{G_t}F_t^l \quad (3)$$

$$B_t = (1-\eta)\ B_{t-1} + \eta \sum_{k=1}^{d} \overline{F_t^k}F_t^k \quad (4)$$

Here $\eta$ is a learning rate. For a new sample $Z$ of size $M\times N$, the maximum response of the target position is,

$$y = F^{-1}\left\{ \frac{\sum_{k=1}^{d}\overline{A^l}Z^l}{B+\lambda} \right\} \quad (5)$$

Furthermore, DSST proposes a fast-scale estimation method to deal with target scale variations. In each frame, the corresponding optimal target scales can be found while estimating target positions. The algorithm first uses a position-dependent filter to determine the new target position. Then, centered on the estimated target position, 33 candidate blocks of different scales are selected, and the optimal matching scale is found by using a scale filter. The image block selection basis is as follows,

$$a^n P \times a^n R, \quad n \in \left\{-\frac{S-1}{2},\ldots,\frac{S-1}{2}\right\} \quad (6)$$

Where $P$ and $R$ indicate the width and height of the target size. S represents the number of scale layers which we set as 33. $a^n$ denotes the scale factors used to obtain different scaled blocks. Then use trained filters to estimate the scale. Finally, the final target state is obtained by combining the results of the position filter and the scale filter.

## 2.2 Algorithm Optimization

There are two main methods of scale estimation. One is the method proposed in DSST and the other is the method proposed in SAMF [8]. DSST uses 33 scale spaces. Then, after extracting the HOG feature for each scale, it is stretched into a one-dimensional vector and a one-dimensional correlation filter is trained online to calculate responses for each scale space. SAMF generates 7 scale spaces based on 7 scale factors. Then, each scale space is interpolated to a fixed size, and the position and scale of the target are calculated. The scale and position of the new target can be determined by searching the peak value of responses.

However, the scale estimation method proposed in DSST consume a large amount of storage resources and computing resources, hence, we adopt the scale estimation method proposed in SAMF. The target response calculation is performed using the same samples as the position estimate in the scale estimation. Firstly, the gray level feature and HOG feature are used to estimate the target position, and then the HOG feature is used to estimate the scale. These two processes are independent of each other, so they can be parallelized, which reduces resource consumption and processing time significantly.

## 2.3 Hardware Architecture

In this paper, we implement the hardware system based on Xilinx XC7K325T development platform. Figure 1 shows the overall architecture of the system.

As shown in Figure 1, the input data must be preprocessed firstly. The image block extraction module extracts the largest-scale image block, and other-scale image blocks are read from the largest-scale image block. The interpolation module inserts the extracted candidate frames of different scales into a fixed size of 128x128. The feature extraction module performs feature extraction on the data in the candidate frame, and the extracted image block size is 32x32.

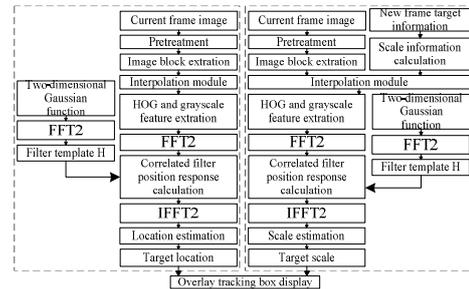

Figure 1. Overall architecture of the hardware system

## 2.4 Implementation Detail

Because the system uses 33 dimensional features (1-D grayscale features and 32-dimensional HOG features). If the filter response of the 33-dimensional feature is simultaneously consumed in parallel, it will consume too much resources. Therefore, the 33-dimensional features are divided into eight batches for processing. In this way, only five layers of related filtering calculation structures need to be designed, which reduces a large part of storage resources and computing resources. The batch processing structure is shown in Fig. 2.

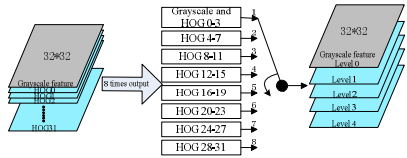

Figure 2. Schematic diagram of batch processing

In this paper, when conducting FFT and IFFT, we need to perform 2D Fourier transform on the input feature map of size 32x32 and 2D inverse Fourier transform on the final result to obtain the final response map. When Fourier transform is performed, one-dimensional Fourier transform is performed on the row data firstly, and the intermediate data is stored and then read out as a column to perform one-dimensional Fourier transform of the column data. Because of their chronological order, the Fourier and inverse Fourier transforms can be optimized. Use an IP core to time-multiplex it to reduce resource consumption. As shown in Figure 3, it shows the schematic diagram before and after optimization：

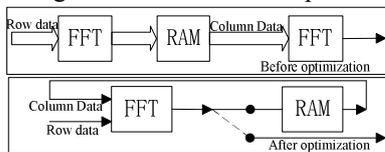

Figure 3. FFT2 calculation structure

## 3. Discussion

We use MATLAB to verify the algorithm. Our algorithm is performed on a PC with intel i7 4790 CPU@3.60 GHz, and the experiment was tested on the OTB50 sub-set. The result as follows. Table 1 is the comparison of our algorithm with other tracker results.

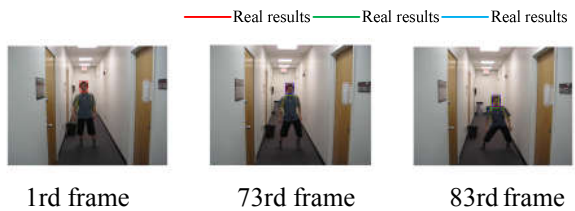

| 1rd frame | 73rd frame | 83rd frame |

Table 1. Tracking speed of our tracker and other trackers

| Method | **OURS** | DSST | Struck | TLD | MIL |
|--------|----------|------|--------|------|-------|
| FPS | **153** | 24 | 5.97 | 22.2 | 22.84 |

It can be seen from table 1, our algorithm has greatly improved accuracy and speed compared to traditional algorithms. Table 2 shows the resource usage of the FPGA.

Table 2. FPGA Resource Utilization Summary

| Resource | Used Resources | Utilization |
|----------|----------------|-------------|
| Slice Registers | 95485 | 23% |
| Slice LUTs | 68433 | 33% |
| Block RAM/FIFO | 179 | 40% |
| DSP48E1s | 143 | 17% |

From the above table, it can be seen that the optimized system structure impressively reduces resource consumption. In the case of less resource consumption, the required functions are well implemented, certain stability and robustness are ensured, the goal of optimizing the hardware implementation is ensured, and the performance of the tracker is significantly improved.

## 4. Summary

In this paper, based on DSST, we improve and optimize the algorithm to meet the hardware implementation requirements for visual tracking. Then we use the FPGA development platform to implement the visual tracking hardware system, and conduct system-level experimentation and verification. The performance of the tracker is evaluated through software and hardware systems. In application, our algorithm has low cost and low power consumption. It can process HD video in real time, with superior robustness and speed. In addition, our algorithm is fast and suitable for low-cost hardware implementation. In the future work, we will further improve the accuracy of the algorithm, so that it can be applied to complex environment.